# Rank Aggregation for Course Sequence Discovery

Mihai Cucuringu
University of California, Los Angeles
mihai@math.ucla.edu

Charlie Marshak
University of California, Los Angeles
cmarshak@math.ucla.edu

Dillon Montag
Westmont College
dmontag23@gmail.com

Puck Rombach
University of California, Los Angeles
rombach@math.ucla.edu

## ABSTRACT

In this work, we adapt the rank aggregation framework for the discovery of optimal course sequences at the university level. Each student provides a partial ranking of the courses taken throughout his or her undergraduate career. We compute pairwise rank comparisons between courses based on the order students typically take them, aggregate the results over the entire student population, and then obtain a proxy for the rank offset between pairs of courses. We extract a global ranking of the courses via several state-of-the art algorithms for ranking with pairwise noisy information, including SerialRank, Rank Centrality, and the recent SyncRank based on the group synchronization problem. We test this application of rank aggregation on 15 years of student data from the Department of Mathematics at the University of California, Los Angeles (UCLA). Furthermore, we experiment with the above approach on different subsets of the student population conditioned on final GPA, and highlight several differences in the obtained rankings that uncover hidden pre-requisites in the Mathematics curriculum.

## 1. INTRODUCTION

College enrollment is at an all-time high at American universities [16], and this generation of college students is choosing to focus on science, technology, engineering, and mathematics (STEM) courses [3]. University faculty and administrators must create systems to effectively and efficiently train this burgeoning population of STEM students. The goal of this paper is to address one aspect of this broad issue through data mining, namely the design of course sequences that can benefit the student population.

For STEM courses, there is often a carefully prescribed sequence of courses to help guide students to the completion of their degree. These prescribed course sequences are designed so that students' core understanding builds on the requisite knowledge and experience for each class they complete, preparing them for the next course. However, the recommended course sequence may not be strictly enforced due to the changing availability of particular courses and the diverse needs of a large, intelligent student population. As a result, students often choose the sequence in which to pursue their courses based on personal preference. This may be based on what course best meets their individual scheduling needs each quarter, or what classes may be more accessible to enroll in at a large university. Students may choose to take a certain course simply based on convenience or because it works toward meeting major requirements. However, they might not take the courses in a sequence that optimally builds their core competency in a subject area. In this work, we apply rank aggregation to obtain course sequences adhered to by UCLA mathematics students to infer hidden dependencies between mathematics courses and to better understand how different types of students navigate their coursework. By comparing the course sequences of A and C students, we are better able to understand optimal course sequences for the mathematics major. In general, we hope to encourage more applications of rank aggregation to order temporal events and discover patterns in sequences.

The remainder of this paper is organized as follows. In Section 2, we review related academic data mining techniques designed for course sequence discovery. In Section 3, we review the technical aspects of the rank aggregation methods that will be used. In Section 4, we apply these methods to analyze sequences of mathematics courses at UCLA. We use student data from the UCLA Department of Mathematics between 2000-2015, and interpret our findings to infer course sequences from these records and latent dependencies between them. We also compare the performance of each rank aggregation method to demonstrate the robustness of this particular framework. In our final Section 5, we review our findings and indicate future directions.

## 2. RELATED WORK

Academic data mining has become a valuable tool for assisting university students in selecting their courses. A popular approach is to adapt eCommerce recommendation systems to the academic space [8, 18, 17]. Here, sequences are determined incrementally by comparing student's records to others with similar coursework and grades. In [19], the authors construct an intricate system that describes how students can move through the web of course dependencies. The authors use their model to extract sequences that minimize the expected time to graduation within their system. Our goal in this work, unlike those above, is to *learn* course sequences by studying the flow of students from course to course. Our approach adapts well-known rank aggregation techniques to extract a complete sequence of all courses based on the partial sequences of courses that students have pursued. In our approach, the extracted course sequence does not provide any indication on how many courses should be taken per quarter nor what courses can be taken simultaneously. However, the extracted sequences can in turn be used as inputs for more personalized recommendation later on. Moreover, these extracted course sequences can be used to understand how different types of students navigate through their major. By comparing high and low performing students, we infer hidden dependencies that might be missed in the probabilistic models.

Rank aggregation has been a powerful tool in web search [6], sport rankings [5, 4], and more recently, grading schemes [14]. However, it has not been used to infer trends in ordering temporal data, and we believe that the data mining of such temporal sequences [21, 13, 20] will benefit from this approach. We now review the state-of-the-art rank aggregation literature, and discuss their applicability to extracting global course sequences that are most consistent with the given data.

## 3. METHODS FOR RANK AGGREGATION

Rank aggregation is the process of obtaining a global ranking from incomplete and noisy pairwise comparisons [6]. Broadly speaking, there are two important steps in such a ranking procedure. First, we translate a pool of inconsistent and incomplete pairwise comparisons into a directed network model, in which links quantify the strength and frequency of these comparisons. Second, we infer global rankings via a variety of spectral methods that consider the weighted adjacency matrix of the network. Such methods are used to rank professional sports teams based on their exhibitions, and in turn, predict the outcomes of elimination tournaments [5, 4]. We will use rankings here not to rank individual players, teams or students, but to determine a global ordering of the courses that is most consistent with the given data. Performing this analysis on students with high versus low GPA has the potential to uncover hidden pre-requisites that could explain and perhaps improve the overall performance of the students with a low final GPA.

For this academic application, we view each student as providing a set of pairwise comparisons between courses based on the order they finished their coursework. These comparisons are often incomplete (students usually do not take all available courses) and noisy (obstacles may force students to take classes out of order). We implicitly assume that there is an underlying, optimal course sequence. We extract this course sequence using rank aggregation methods, which we describe in detail in Section 3. A review of many of the methods used here can be found in [5]. We apply six different methods to extract course sequences: PageRank [12], Rank Centrality [11], SerialRank [7], SyncRank [5], SVD-Rank [5], and a Least-Squares based method. We first describe the network models that we use for all rank aggregation methods.

### 3.1 Network Model and Ranking Constructions

For our academic application of rank aggregation, we wish to extract a global ranking of all courses that is most consistent with the order in which they are taken by students. Here, our comparisons stem from the frequency with which course $i$ was taken before course $j$. The first step to extract course sequences will be to translate these comparisons into a network model. We present two network models, each quantifying the flow of students from course to course.

In the first network model, each node represents a course, and our edges represent the flow of students between two courses. Let $k = 1, \ldots, n_s$ be our enumeration of students and similarly $i, j$ be courses with $i, j = 1, \ldots, n_c$. We first define the variable $I_{ij}^k$ as the indicator that student $k$ took course $i$ before course $j$

$$I_{ij}^k = \begin{cases} 1 & \text{if student } k \text{ took course } i \text{ before } j \\ 0 & \text{otherwise.} \end{cases}$$

The count matrix $\mathbf{C}$ is defined as

$$C_{ij} = \sum_{k=1}^{n_s} I_{ij}^k \quad \text{when } i \neq j,$$

and $C_{ij} = 0$ when $i = j$. We define the $n_c \times n_c$ transition matrix $\mathbf{P}$ to have entries representing the number of students that have taken course $i$ before $j$, as a percentage of all students that have taken both course $i$ and $j$,

$$P_{ij} = \frac{C_{ij}}{C_{ij} + C_{ji}}.$$

If $C_{ij} = C_{ji} = 0$, then we let $P_{ij} = P_{ji} = 0$. By construction, $P_{ij} + P_{ji} = 1$ when course $i$ and course $j$ are compared at least once. The matrix $\mathbf{P}$ defines a directed multigraph in which an edge weight $P_{ij}$ represents the flow of students moving from course $i$ to course $j$. In Figure 1, we illustrate a portion of the network model on three courses: Discrete Structures, Linear Algebra I, and Real Analysis I. Here, we consider only Applied Mathematics students. We note that of those students that take Real Analysis I and Linear Algebra I in different quarters, very few take Real Analysis first.

We now define a second related network model in which we ensure the net flow of students between two courses is 0, which is necessary for some of the ranking methods. We define the skew-symmetric matrix $\mathbf{F}$ of size $n_c \times n_c$, with $|F_{ij}| \in [0.5, 1]$, which encodes the frequency with which course $i$ is taken before course $j$. We define $F_{ij}$ using $P_{ij}$'s above as

$$F_{ij} = \begin{cases} P_{ij} & \text{if } P_{ij} \geq 0.5 \\ P_{ij} - 1 & \text{if } P_{ij} < 0.5. \end{cases} \quad (1)$$

For example, if 70% of the students (who took both courses $i$ and $j$) took course $i$ before $j$, then we set $F_{ij} = 0.7$ and $F_{ji} = -0.7$.

In larger scale applications, the measurement matrices $\mathbf{F}$ and $\mathbf{P}$ will most likely be sparse, with only a small subset of the possible pairwise comparisons present. Let $G(V, E)$ be a graph, where the node set $V$ represents the set of courses, with $|V| = n_c$. We add an edge between course $i$ and $j$, that is $(i,j) \in E$ whenever $P_{ij} \neq 0$ (equivalently, $F_{ij} \neq 0$), and let $m = |E|$ be the number of edges in $G$.

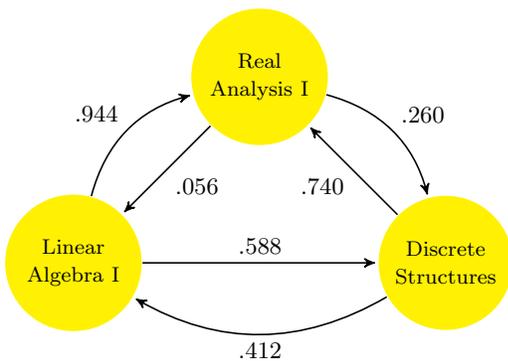

**Figure 1: A subgraph corresponding to P for Applied Mathematics majors. The courses shown are Linear Algebra I, Real Analysis I, and Discrete Structures).**

### 3.2 PageRank

PageRank was designed to "bring order to the web", ranking sites according to the distribution of links between them [12]. In [4], the authors adapted PageRank to determine NCAA basketball team rankings based on the regular season matchups [4]. This in turn allowed the authors to forecast the outcome in the NCAA tournament based on the rankings they obtained. This adaptation is a form of rank aggregation using pairwise comparisons that are either incomplete (not all teams play each other) or noisy (teams may under or over perform). Using the network determined by $\mathbf{P}$ (section 3.1), PageRank defines a stochastic matrix $\mathbf{S}_\alpha$ that describes the motion of a random student who is permitted to teleport to non-adjacent nodes [9]. This stochastic matrix $\mathbf{S}_\alpha$ is given by

$$\mathbf{S}_\alpha^{\mathrm{pr}} = (1-\alpha)\mathbf{D}^{-1}\mathbf{P} + \frac{\alpha}{n_c}\mathbf{I}_{n_c},$$

where $\alpha \in (0,1)$, $\mathbf{I}_{n_c}$ is the $n_c \times n_c$ identity matrix, and $\mathbf{D}$ is the diagonal matrix of out degrees with $d_{ii} = \sum_j P_{ij}$. The first term $(1-\alpha)\mathbf{D}^{-1}\mathbf{P}$ means that with probability $(1-\alpha)$ the random walker travels to an adjacent node in the network. The second term $\frac{\alpha}{n_c}\mathbf{I}_{n_c}$ means that with probability $\alpha$ the random walker moves to any other node in the network. The teleportation in this setting can be interpreted as the need for a student to move randomly in the course sequence to fulfill a requirement.

Formally, we associate the Markov chain with states $\mathbf{q}_t$ at time $t$, and with the transition rule given by

$$(\mathbf{q}_{t+1})^T = (\mathbf{q}_t)^T \mathbf{S}_\alpha^{\mathrm{pr}}.$$

The PageRank vector $\mathbf{q}_\infty$ is defined to be the stationary distribution associated to the above Markov chain and given as $\lim_{t\to\infty} \mathbf{q}_t$. We determine the vector $\mathbf{q}_\infty$ via power iteration [9]. The $i$th component of $\mathbf{q}_\infty$ denotes the ranking of course $i$. The smaller the entry at $i$, the more likely students take this course early in their sequence. From this ordering we extract a global ranking for courses. Using this stationary distribution, we adapt personalized PageRank [9] replacing $\mathbf{I}_{n_c}$ with $\mathbf{1}_{n_c} - \mathbf{q}_\infty$, where $\mathbf{1}_{n_c}$ is the vector of length $n_c$ of 1s. The intuition is that students teleport to fulfill requirements early in their course sequence before moving on.

### 3.3 Rank Centrality

Rank Centrality was conceived as a way to discover rankings generated by the Bradley-Terry model [11]. This model assumes players $i$ and $j$ have latent real-valued weights $w_i, w_j$ assigned to them so that

$$\mathbb{P}(i \text{ beats } j) = \frac{w_i}{w_i + w_j}.$$

The authors used this method to in turn rank NASCAR drivers and Indian cricket teams [11]. Again, for our application, we compare courses $i$ and $j$, in which "beat" means course $i$ came before course $j$ in sequence of courses. Let $\mathbf{P}$ be the matrix defined in the previous section whose entries $P_{ij}$ denote the number of students that took course $i$ before $j$, amongst all those students that took $i$ and $j$ in different quarters. Rank Centrality defines a Markov chain on the $n_c$ courses with the following stochastic matrix

$$\mathbf{S}^{\mathrm{rc}} = \frac{1}{d_{\max}}\mathbf{P} + \left(\mathbf{I}_{n_c} - \frac{1}{d_{\max}}\mathbf{D}\right),$$

where $d_{\max}$ is the maximum out degree in the network, $\mathbf{I}_{n_c}$ is the $n_c \times n_c$ identity matrix, and $\mathbf{D}$ is the diagonal matrix of out degrees with $d_{ii} = \sum_j P_{ij}$. This differs in two important ways from PageRank above. First, there is no teleportation term for Rank Centrality. Second, from the latter term in the sum, the random walker can remain at course $i$ with probablity $1 - 1/d_{\max}\sum_j P_{ij}$. It also means that courses with smaller total out-degree will have an added self-loop of nontrivial weight. This means less-popular courses tend to have higher marginal values in the stationary distribution.

The above construction has useful theoretical properties. For one, this stochastic process is reversible. Moreover, if we assume $\mathbf{P}$ is a realization of a Bradley-Terry model, the stationary distribution will be proportional to the weight vector $\mathbf{w}^T = (w_1, ..., w_{n_c})$. We compute the stationary distribution $\mathbf{q}_\infty = \lim_{\to\infty} \mathbf{q}_t \mathbf{S}^{\mathrm{rc}}$ using power iteration.

### 3.4 SerialRank

SerialRank [7] adapts the seriation problem proposed in [2] to determine a global ranking of players. In SerialRank, the authors define a similarity function determined by the outcomes two players have with common opponents. As such, they study the similarity graph rather than a Markov chain, as is done in the previous two rank aggregation methods. For our academic application, we interpret links as the likelihood two courses are taken at a similar time.

To construct the similarity matrix, we recall the matrix $\mathbf{P}$ from the previous sections whose entries $P_{ij}$ that counts the percentage of students taking course $i$ before $j$. We proceed as in [7]. We construct the comparison matrix $\mathbf{A}_k$ for course $k$ to be given component-wise by

$$(A_k)_{ij} = 1 - \frac{|P_{ik} - P_{jk}|}{2}$$

whenever both course $i$ and course $j$ have been taken in a sequence with course $k$. If either course $i$ or course $j$ has not been taken in sequence $k$, we define $(A_k)_{ij} = \frac{1}{2}$. In other words, if course $i$ and course $j$ have a similar percentage of students that are taking course $k$ afterwards, then $i$ and $j$ must be more similar themselves. The similarity matrix

$\mathbf{S}^{\mathrm{sr}}$ is then determined by summing over the comparison matrices

$$\mathbf{S}^{\mathrm{sr}} = \sum_{k=1}^{n_c} \mathbf{A}_k.$$

We rescale $\mathbf{S}^{\mathrm{sr}}$ by subtracting the minimum value of $\mathbf{S}^{\mathrm{sr}}$ so that course $i$ and $j$ with the smallest similarity now have similarity of 0. To determine a ranking from $\mathbf{S}^{\mathrm{sr}}$, we form the combinatorial laplacian $\mathbf{L}$ and rank the courses using the components of the Fiedler vector [7], where $\mathbf{L}$ is given by

$$\mathbf{L} = \mathbf{D} - \mathbf{S}^{\mathrm{sr}}.$$

The justification for using the Fiedler vector is proposed in [2] as a relaxation of an NP-hard problem. Specifically, let us assume that $\mathbf{q}$ is a vector with the $i$th component representing the ranking of the $i$th course. We can see that the following energy will be minimized for an optimal ranking

$$\mathbf{q} = \arg\min_{\mathbf{q}'} \sum_{i,j} s_{ij}^{\mathrm{sr}} (q_i' - q_j')^2.$$

Observing that $\sum_{i,j} s_{ij}^{\mathrm{sr}} (q_i' - q_j')^2 = (\mathbf{q}')^T \mathbf{L} \mathbf{q}'$, the problem can be relaxed into the well-studied eigenvector problem

$$\mathbf{q} = \arg\min_{\mathbf{q}'} (\mathbf{q}')^T \mathbf{L} \mathbf{q}'$$
$$\text{such that} \quad ||\mathbf{q}'||_2 = 1 \text{ and } \mathbf{1}_{n_c}^T \mathbf{q}' = 0,$$

where $\mathbf{1}_{n_c}$ is the vector of length $n_c$ comprised entirely of 1's. The minimum of the above optimization problem is attained by the eigenvector corresponding to the smallest non-trivial eigenvalue of $\mathbf{L}$. After ordering the $n_c$ courses using the components of $\mathbf{q}$, we obtain a global ranking. Theoretical results in [7, 2] give guarantees on the monotonicity of $\mathbf{q}$ for particular generative models.

### 3.5 SyncRank: Synchronization based ranking

In [5], one of the authors formulates the problem of ranking with incomplete noisy information as an instance of the group synchronization problem over the special orthogonal group SO(2), which we briefly detail. Determining individual group elements from the measurement of their pairwise ratios is known as the *group synchronization* problem [15]. The seminal paper of Singer [15] considered the angular synchronization problem over SO(2), where the goal is to recover $n$ unknown ground truth angles $\theta_1, \ldots, \theta_n \in [0, 2\pi)$, given $m$ noisy pairwise angle offsets captured in a matrix $\Theta$ of size $n \times n$

$$\Theta_{ij} = (\theta_i - \theta_j + \text{Noise}) \mod 2\pi.$$

Singer [15] introduced and analyzed spectral and semidefinite programming (SDP) relaxations for this problem (followed by a rounding procedure). The difficulty of the problem is amplified on one hand by the amount of noise in the offset measurements, and on the other hand by the fact that $m \ll \binom{n}{2}$, i.e., only a very small subset of all possible pairwise offsets are measured.

The connection between ranking and angular synchronization can be summarized as follows. Denote the true ranking of the $n_c$ courses by $r_i$, assuming, without loss of generality, that $r_i = i$, *i.e.*, the rank of the $i^{th}$ course is $i$. Imagine the ranks to lie on a one-dimensional line, with the pairwise rank comparisons given, in the noiseless case, by $R_{ij} = r_i - r_j$.

Note that this matrix has a similar interpretation to the matrix $F$ defined in (1), in the sense that both are skew-symmetric matrices that capture the outcome of the pairwise comparison between $i$ and $j$.

Next, we consider the angular embedding given by mapping the ranks of the courses to the unit circle, say fixing $r_1$ to have a zero angle with the $x$-axis, and the last player $r_{n_c}$ corresponding to an angle equal to $\pi$. We interpret the given pairwise measurements $F_{ij}$ as a proxy for the rank offsets, and map them to $\Theta_{ij} \in [0, \pi)$ via the transformation

$$R_{ij} \mapsto \Theta_{ij} := \pi F_{ij}. \tag{2}$$

In other words, we imagine the $n_c$ courses wrapped around a fraction of the circle, interpret the available rank-offset measurements as angle-offsets in the angular space, and thus arrive at the setup of the angular synchronization problem. We then build the $n_c \times n_c$ Hermitian matrix $\mathbf{H}$ with $H_{ij} = e^{\iota \theta_{ij}}$, if $(i, j) \in E$, and $H_{ij} = 0$ otherwise, with $\iota = \sqrt{-1}$. We solve the angular synchronization problem via its spectral relaxation, which considers the top eigenvector $v_1$ of $\mathbf{H}$, and denote the recovered solution by $\widehat{r}_i = e^{\iota \widehat{\theta}_i} = \frac{v_1(i)}{|v_1(i)|}$, $i = 1, 2, \ldots, n_c$. We extract the corresponding set of angles $\widehat{\theta}_1, \ldots, \widehat{\theta}_{n_c}$ from $\widehat{r}_1, \ldots, \widehat{r}_{n_c}$, which induces the final ranking solution after modding out the best circular permutation. This last step is due to the fact that the estimation of the rotation angles is up to an additive phase, since $e^{\iota \alpha} v_1$ is also an eigenvector of $\mathbf{H}$ for any $\alpha \in \mathbb{R}$.

### 3.6 Least Squares

Another method we consider for recovering the course orderings is one based on the traditional least-squares approach. Denote by $m$ the number of edges in the measurement graph $G$, $m = |E(G)|$, which counts the number of available pairwise comparisons, and by $\mathbf{B}$ the edge-vertex incidence matrix of size $m \times n_c$, with entries given by

$$B_{ij} = \begin{cases} 1 & \text{if } (i,j) \in E(G), \text{ and } i > j \\ -1 & \text{if } (i,j) \in E(G), \text{ and } i < j \end{cases} \tag{3}$$

We let $\mathbf{w}$ the vector of length $m \times 1$ which contains the pairwise rank measurements $w_e = F_{ij}$, for all edges $e = (i, j) \in E(G)$. We consider the resulting over-determined system of linear equations, and compute the least-squares solution

$$\underset{\mathbf{x} \in \mathbb{R}^n}{\text{minimize}} ||\mathbf{B}\mathbf{x} - \mathbf{w}||_2^2. \tag{4}$$

Hirani *et al* [10] show that the problem of least-squares ranking on graphs has far-reaching rich connections with various other research areas, including spectral graph theory and multilevel methods for graph Laplacian systems, Hodge decomposition theory and random clique complexes in topology.

### 3.7 Ranking via Singular Value Decomposition

The sixth and final ranking method we consider is the recent SVD-based approach proposed in recent work by one of the authors [5], and based on the traditional Singular Value Decomposition (SVD). Let $\mathbf{r}$ be the ranking vector with $r_i$ denote the rank of the $i$th player. The applicability of the SVD-Rank approach stems from the observation that, the noiseless matrix of rank offsets defined as before to be

$R_{ij} := r_i - r_j$ is a skew-symmetric matrix of rank 2 and

$$\mathbf{R} = \mathbf{r}\mathbf{e}^T - \mathbf{e}\mathbf{r}^T, \quad (5)$$

where $\mathbf{e}$ denotes the all-ones column vector.

For a certain noise model, $\mathbf{R}$ can be shown to decompose into a low-rank (rank-2) perturbation of a random skew-symmetric matrix [5]. In practice, we are not given access to the clean matrix $\mathbf{R}$ of rank offsets, and use as a proxy the available matrix $\mathbf{F}$ given by (1). We consider the top two singular vectors of $\mathbf{F}$, order their entries by size, extract the resulting rankings, and choose between the first and second singular vector based on whichever one yields a better consistency coefficient, which we will define formally in the next section (6). Note that since the singular vectors are obtained up to a global sign, we (again) choose the ordering which is most consistent with the given data, by comparing the resulting consistency coefficients (6).

## 4. COURSE SEQUENCE DISCOVERY

We now apply the six methods discussed in the previous section to data collected by the UCLA Department of Mathematics between Fall 2000 and Spring 2015. We first compare the output of each method on all of these student records and then demonstrate the applicability of this framework by quantifying the flow of students through an obtained sequence. After assessing each method, we analyze course sequences of particular student populations to highlight a few trends. In particular, by comparing the course sequences of high and low performing students, we infer that there are hidden prerequisites. This means that some courses should follow others, so that students are building on their core competency based on a specific sequence of courses. We now briefly discuss the data itself and the cleaning of it.

### 4.1 Cleaning the Data

The data is comprised of individual student records indicating each student's quarter by quarter math coursework, the student's grade in the course, and the major declared each quarter. To apply the rank aggregation methods above, we must construct the matrix $\mathbf{P}$ or $\mathbf{F}$ defined in Section 3.1. We restrict our attention to a particular major, each major having their very own set of requirements and objectives. At UCLA, the Department of Mathematics has 7 unique majors ranging from Pure Mathematics to Mathematics & Economics to Mathematics of Computation. Since a student can change their major from quarter to quarter, we group students based on the major they declare in their last quarter. We also exclude all community college transfer students as they often exhibit very different trends compared to students that were admitted as Freshman. Moreover, if a student retook a class, we look only at the grade and the quarter the class was *last* taken. Lastly, once a suitable population of students is selected, we only consider those math classes that at least 10 percent of this population takes. In particular, if we consider only a particular major, we exclude classes that less than 10% of students in this major has taken.

### 4.2 Assessing Rank Aggregation

We assess our rankings quantitatively by using a metric, and qualitatively by examining the output sequences. To the first end, we define an consistency coefficient that roughly measures both how well-ordered a course sequence is and how well the ranking captures this ordering. Clearly, there is no course sequence that all students follow perfectly, but we do expect certain courses in aggregate to follow a sequence. We assess our rankings measuring the flow of students from course to course when we follow a given sequence. We consider the matrix $\mathbf{P}$ defined in (3.1). If two courses $i$ and $j$ are independent and offered with similar frequency, we expect $P_{ij} \approx .5$. However, when course $i$ crucially depends on course $j$, we expect $P_{ij} > .5$. Let $\mathbf{r}$ be a ranking of the $n_c$ courses. Without loss of generality, we assume $r_i = i$. We define the consistency coefficient $\Gamma(\mathbf{r})$ as in [5]:

$$\Gamma(\mathbf{r}) = \frac{4}{n^2 - n} \sum_{i<j} (P_{ij} - .5). \quad (6)$$

The coefficient $\Gamma(\mathbf{r})$ is normalized so that $\Gamma(\mathbf{r}) \in [-1, 1]$. Note that in the event that all the courses are taken in random order by students, we expect $\Gamma(\mathbf{r}) \approx 0$ for any ranking $\mathbf{r}$. On the other hand, if $\Gamma(\mathbf{r})$ is close to 1 for some ranking, we infer that $\mathbf{r}$ was suitable for discovering an underlying course sequence that many students followed. In other words, the closer $\Gamma(\mathbf{r})$ is to 1, the more consistent the given data is with the existence of a global ordering. There are other ways to assess the strength of rankings as in [5] and by no means do we claim $\Gamma(\mathbf{r})$ a definitive assessment. Certainly, this metric neither takes into account how many students took $i$ and $j$, nor does it consider the distance typical between courses $i$ and $j$. In Table 1, we provide values of $\Gamma(\mathbf{r})$ for all six methods and their corresponding rankings $\mathbf{r}$ for three different majors across four different performance categories. The majors considered are Applied Mathematics, Applied

**Table 1:** $\Gamma(\mathbf{r})$ coefficients for Two Majors.

| Method | Applied Mathematics ($n_s = 672$) | | | |
|---|---|---|---|---|
| | All | A | B | C |
| PageRank | 0.652 | 0.691 | 0.669 | 0.674 |
| Rank Centrality | 0.531 | 0.671 | 0.546 | 0.670 |
| SerialRank | 0.657 | 0.696 | 0.674 | 0.678 |
| SyncRank | 0.657 | 0.696 | 0.669 | 0.679 |
| Least Squares | 0.654 | 0.697 | 0.676 | 0.679 |
| SVD | 0.657 | 0.701 | 0.674 | 0.678 |
| | Applied Science ($n_s = 499$) | | | |
| | All | A | B | C |
| PageRank | 0.707 | 0.772 | 0.722 | 0.757 |
| Rank Centrality | 0.727 | 0.743 | 0.740 | 0.757 |
| SerialRank | 0.725 | 0.768 | 0.737 | 0.767 |
| SyncRank | 0.725 | 0.763 | 0.737 | 0.768 |
| Least Squares | 0.725 | 0.760 | 0.727 | 0.771 |
| SVD | 0.715 | 0.746 | 0.735 | 0.766 |
| | Pure Mathematics ($n_s = 346$) | | | |
| | All | A | B | C |
| PageRank | 0.618 | 0.706 | 0.634 | 0.645 |
| Rank Centrality | 0.606 | 0.708 | 0.622 | 0.685 |
| SerialRank | 0.617 | 0.713 | 0.631 | 0.692 |
| SyncRank | 0.617 | 0.721 | 0.632 | 0.694 |
| Least Squares | 0.619 | 0.718 | 0.634 | 0.707 |
| SVD | 0.618 | 0.714 | 0.632 | 0.692 |

Science and Pure Mathematics. The first two majors are the largest in all of mathematics and we expect there to be the most noise. Within each major, we illustrate three GPA categories. From the table, we see that in general, $\Gamma(\mathbf{r})$ is maximal for A range students and minimized when we consider the entire major population. The large value of $\Gamma(\mathbf{r})$ for A students may be an artifact of this population being substantially smaller and perhaps more homogeneous in terms of the ordering in which classes were taken. We also note that Rank Centrality performs rather poorly relative to the other methods for Applied Mathematics majors, and offer below a possible explanation for this. Despite this one particular under performance of Rank Centrality, the coefficients for $\Gamma(\mathbf{r})$ are all within .015 of each other, an agreement which demonstrates that for the small data set at hand, all the ranking methods perform rather similarly.

We can also analyze the sequences obtained by rank aggregation. In Table 2, we compare the output sequences for all six methods on Applied Mathematics majors who have an A-range GPA. All methods were able to correctly order the calculus sequences (31AB, 32AB, 33AB) and are thus removed from the table. Moreover, all methods place Linear Algebra I and Real Analysis I fairly early in the sequence as these are courses required for all majors. All the methods consistently placed Real Analysis I before Real Analysis II and III and similarly for other courses taught in sequences. These basic dependencies were captured by rank aggregation across the board. We did notice that Rank Centrality had the greatest disagreement when compared to the rest of the methods. Rank Centrality placed Applied Algebra and Partial Differential Equations much earlier than the other methods. We suspect the defect in $\Gamma(\mathbf{r})$ seen in Table 1 and this particular disagreement with the other methods is due to the relatively low enrollment in these classes relative to those courses considered. Approximately 10% of students in the Applied Mathematics major took Partial Differential Equations and Applied Algebra, which was the minimum for a class to be considered in our course sequence. We suspect Rank Centrality to be affected in certain cases by the popularity of classes. We also display the heat-map corresponding to the matrix of $\mathbf{P}$ in Figure 2 to reflect a more granulated view of the flow of students. See Appendix A for a reference of all the course numbers and names that label the axes. Here, we only consider Pure Math students with A-range GPAs. Again, we can see from the same Figure 2 that the Calculus sequence (31A, 31B, 32A, 32B) is taken earliest in the expected sequence. We observe $P_{ij} + P_{ji} = 1$ when both $i$ and $j$ have been taken by at least one student and are taken in different quarts. There are some pairs of classes for which this is not the case. For instance, we can see from Figure 2 that no single student took Math 61 (Discrete Structures) and Math 133 (Fourier Series) in different quarters. Our assessments above indicate that indeed this approach to course sequence discovery produces reasonable results.

## 4.3 Inferring Hidden Prerequisites

One of the original motivations for this work was to learn optimal course sequences through the math major. Even though our network model is not dependent on performance, we can still condition the students we consider in the network's construction by their overall GPA. Comparing the extracted sequences of high and low performing students, we can infer hidden prerequisites and optimal course sequences. In Table 3, we do just this for three different majors using SyncRank [5]. We compare the first 11 non-Calculus courses for A-range students and C-range students. For pure mathematics students, we see that A students tend to take Discrete Structures before Linear Algebra I, the latter course being the first upper division proof-based course a math student takes at UCLA. As such, we might infer that Discrete Structures, which covers the basics of functions, sets, and combinatorics to be helpful to Pure Mathematics students entering a proof-based curriculum. Another interesting feature for Pure Mathematics students is that A students tend to take Real Analysis I and Real Analysis II fairly close to each other while C students do not. The Real Analysis coursework is known to be a conceptually challenging university course, both at UCLA and elsewhere. More statistical analysis is needed to ascertain whether strong performance is correlated to taking these course sequences in close succession, but rank aggregation allows us to quickly identify such trends.

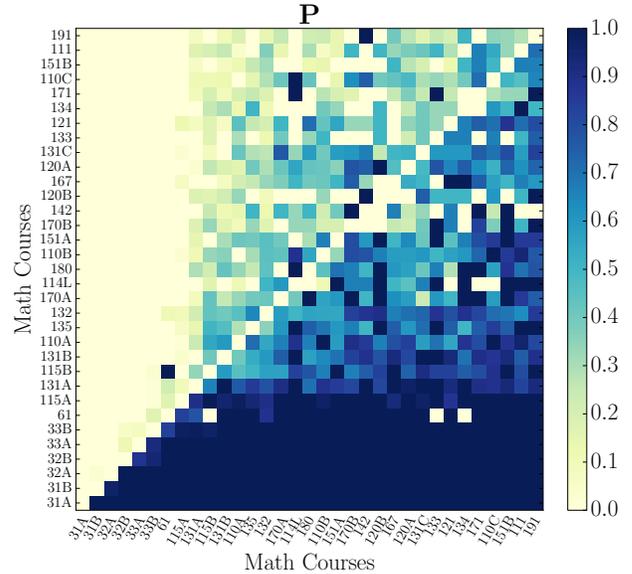

**Figure 2:** Above, the P matrix for A students with a Pure Mathematics focus. The courses are ordered by PageRank.

For Applied Mathematics students, we see that Probability I is frequently taken after Real Analysis I for A students, but not for C students. As such, we infer that Real Analysis I is a prerequisite for Probability I when considering such students. Contrary to this ordering, Applied Science students take Probability I before Real Analysis I whether they are A or C students. One possible explanation for the differences between these applied majors is that many Applied Science students are pre-actuarial and are very familiar with the material in Probability I. As such, the ordering of Probability I and Real Analysis I is not as pertinent for Applied Science students.

Assuming A students navigate through the major best, we can also use rank aggregations method to study differences in major course sequences as in Table 4. Here, we apply SerialRank to 6 different math majors, again only considering A-range students. We see that the Applied math disciplines (Applied Science, Mathematics & Economics, Mathematics for Computation, and Applied Mathematics) all take Probability I early, while Pure Mathematics students generally do not. We also noticed that Applied Science and Mathematics for Computation students take at least one pre-actuarial course, while other majors do not. Of course, these findings do not indicate the correlation of these course sequences with performance, but this methodology quickly illuminates possible trends.

Table 2: First 11 courses for A students in 5 different majors using SerialRank.

| Applied Mathematics $n_s = 140$ | Pure Mathematics $n_s = 86$ | Applied Sciences $n_s = 75$ | Mathematics & Economics $n_s = 101$ | Mathematics for Computation $n_s = 20$ |
| --- | --- | --- | --- | --- |
| Lin. Algebra I | Discr. Struct. | Lin. Algebra I | Lin. Algebra I | Lin. Algebra I |
| Discr. Struct. | Lin. Algebra I | Probability I | Discr. Struct. | Probability I |
| Real Analysis I | Real Analysis I | Discr. Struct. | Probability I | Real Analysis I |
| Probability I | Lin. Algebra II | Act. Math | Real Analysis I | Discr. Struct. |
| NonLin. Systems | Real Analysis II | Probability II | Applied Algebra | Probability II |
| Num. Analysis I | Algebra I | Real Analysis I | Optimization | Act. Math |
| Graph Theory | Ord. Diff. Eqn.'s | Num. Analysis I | Probability II | Math Modeling |
| Complex Analysis | Complex Analysis | Act. Models II | Num. Analysis I | Math Econ. |
| Real Analysis II | Probability I | Graph Theory | Real Analysis II | Act. Models II |
| Math Modeling | Algebra II | Optimization | Game Theory | Num. Analysis I |
| Algebra I | Graph Theory | Act. Models II | Math Econ. | Loss Models I |

Table 3: Comparing the A and C students in 3 majors using SyncRank.

| Applied Mathematics | | Applied Sciecnces | | Pure Mathematics | |
| --- | --- | --- | --- | --- | --- |
| A ($n_s = 140$) | C ($n_s = 198$) | A ($n_s = 75$) | C ($n_s = 162$) | A ($n_s = 86$) | C ($n_s = 95$) |
| Lin. Algebra I | Lin. Algebra I | Lin. Algebra I | Lin. Algebra I | Discr. Struct. | Lin. Algebra I |
| Discr. Struct. | Discr. Struct. | Probability I | Discr. Struct. | Lin. Algebra I | Hist. of Math |
| Real Analysis I | Probability I | Discr. Struct. | Probability I | Real Analysis I | Real Analysis I |
| Probability I | Real Analysis I | Real Analysis I | Real Analysis I | Lin. Algebra II | Discr. Struct. |
| Complex Analysis | Algebra I | Act. Math | Nonlin. Syst. | Algebra I | Algebra I |
| Nonlin. Syst. | Num. Analysis I | Num. Analysis I | Math Modeling | Real Analysis II | Ord. Diff. Eqn.'s |
| Num. Analysis I | Graph Theory | Probability II | Graph Theory | Ord. Diff. Eqn.'s | Complex Analysis |
| Math Modeling | Real Analysis II | Graph Theory | Game Theory | Complex Analysis | Game Theory |
| Real Analysis II | Act. Math | Act. Models II | Num. Analysis I | Probability I | Probability I |
| Algebra I | Nonlin. Syst. | Act. Models II | Optimization | Algebra II | Graph Theory |
| Graph Theory | Math Modeling | Ord. Diff. Eqn.'s | Ord. Diff. Eqn.'s | Graph Theory | Num. Analysis I |
| Ord. Diff. Eqn.'s | Hist. of Math | Num. Analysis II | Act. Math | Real Analysis III | Optimization |
| Game Theory | Complex Analysis | Optimization | Probability II | Num. Analysis I | Number Theory |
| Research Seminar | Probability II | Math Econ. | Act. Models II | Logic | Algebra II |

## 5. CONCLUSIONS AND FUTURE WORK

In this paper, we demonstrated that with an appropriate network model, rank aggregation techniques can extract course sequences and infer latent course dependencies. Our findings captured easily verifiable dependencies such as the completion of lower division calculus courses, and then proceeding to Linear Algebra I and Real Analysis I, two gateway classes central to the math curriculum across all majors. We were able to inspect the differences between the various math majors at UCLA and the differences between A, B, and C students within these majors. We applied six different methods of rank aggregation: PageRank, Rank Centrality, SerialRank, SyncRank, Least Squares, and SVD ranking. We then compared these methods using an consistency coefficient $\Gamma(\mathbf{r})$ determined by rankings defined in Eq. (6). We concluded that the output of all the ranking methods considered were in rather close agreement except for Rank Centrality, for which the ranking of a course was affected by low enrollment. Lastly, we were able to infer that there are some hidden prerequisites of courses that emerge based on the trends observed in high and low performing students.

Rank aggregation has typically been used to rank sports teams, athletes, or other competitive endeavors. The application in our present work adapts this methodology for temporal orderings of university course data. The crucial ingredient was an appropriate network model that captured how students navigate through courses, which rendered the problem suitable for existing state-of-the-art algorithms for ranking with incomplete and inconsistent pairwise measurements.

## 6. ACKNOWLEDGEMENTS

This work was supported by NSF grant DMS-1045536, UC Lab Fees Research Grant 12-LR-236660, ARO MURI grant W911NF-11-1-0332, AFOSR MURI grant FA9550-10-1-0569,NSF grant DMS-1417674, and ONR grant N-0001-4121-0838. We were grateful to Andrea Bertozzi and Dimitri Shlyakhtenko for accruing this academic data for research and their careful oversight throughout the project. Many of the data mining experiments were performed during the 2015 UCLA summer Applied Mathematics Research Experience for Undergraduates (REU). As such, we would like to acknowledge the invaluable conversations with Juan Carlos Apitz, Jessica Tran, Ritvik Kharkar, and Milicia Hadzi-Tanovic, with whom we worked extensively on pre-processing the data and understanding student trends. Finally, we would also like to thank Fajwel Fogel for helpful discussions regarding SerialRank.

Table 4: Course Sequences for Applied Mathematics Majors obtained by the six methods.

| PageRank | Rank Centrality | SerialRank | SVD | Least Squares | SyncRank |
|---|---|---|---|---|---|
| Linear Algebra I | Linear Algebra I | Linear Algebra I | Linear Algebra I | Linear Algebra I | Linear Algebra I |
| Discr. Struct. | Discr. Struct. | Discr. Struct. | Discr. Struct. | Discr. Struct. | Discr. Struct. |
| Real Analysis I | Applied Algebra | Real Analysis I | Real Analysis I | Real Analysis I | Real Analysis I |
| Probability I | Partial Diff. Eqn.'s | Probability I | Probability I | Probability I | Probability I |
| Num. Analysis I | Math Modeling | Num. Analysis I | Nonlin. Syst. | Num. Analysis I | Num. Analysis I |
| Nonlin. Syst. | Real Analysis I | Nonlin. Syst. | Num. Analysis I | Nonlin. Syst. | Nonlin. Syst. |
| Complex Analysis | Abstract Algebra I | Abstract Algebra I | Complex Analysis | Complex Analysis | Complex Analysis |
| Abstract Algebra I | Game Theory | Complex Analysis | Abstract Algebra I | Abstract Algebra I | Abstract Algebra I |
| Real Analysis II | Complex Analysis | Real Analysis II | Real Analysis II | Real Analysis II | Act. Math |
| Graph Theory | Num. Analysis II | Graph Theory | Graph Theory | Graph Theory | Graph Theory |
| Math Modeling | Num. Analysis I | Math Modeling | Act. Math | Math Modeling | Real Analysis II |
| Act. Math | Graph Theory | Act. Math | Math Modeling | Act. Math | Applied Algebra |
| Ord. Diff. Eqn.'s | Probability I | Ord. Diff. Eqn.'s | Applied Algebra | Applied Algebra | Math Modeling |
| Applied Algebra | Nonlin. Syst. | Applied Algebra | Ord. Diff. Eqn.'s | Ord. Diff. Eqn.'s | Ord. Diff. Eqn.'s |
| Optimization | Hist. of Math | Hist. of Math | Hist. of Math | Hist. of Math | Hist. of Math |
| Hist. of Math | Probability II | Probability II | Probability II | Optimization | Probability II |
| Probability II | Act. Math | Optimization | Optimization | Probability II | Optimization |
| Num. Analysis II | Real Analysis II | Game Theory | Num. Analysis II | Num. Analysis II | Num. Analysis II |
| Game Theory | Ord. Diff. Eqn.'s | Num. Analysis II | Game Theory | Game Theory | Game Theory |
| Partial Diff. Eqn.'s | Optimization | Partial Diff. Eqn.'s | Partial Diff. Eqn.'s | Partial Diff. Eqn.'s | Partial Diff. Eqn.'s |

Table 5: Course names and numbers.

| # | Course Name | # | Course Name | # | Course Name | # | Course Name |
|---|---|---|---|---|---|---|---|
| 31A | Calculus I | 111 | Number Theory | 132 | Complex Analysis | 170B | Probability II |
| 31B | Calculus II | 115A | Linear Algebra I | 133 | Fourier Analysis | 172A | Actuarial Mathematics |
| 32A | Multivariable Calculus I | 115B | Linear Algebra II | 134 | Nonlinear Systems | 172B | Actuarial Models II |
| 32B | Multivariable Calculus II | 117 | Applied Algebra | 135 | Ordinary Differential Equations | 172C | Actuarial Models II |
| 33A | Linear Algebra for Applications | 120A | Differential Geometry I | 136 | Partial Differential Equations | 173A | Casualty Loss Models I |
| 33B | Differential Equations for Applications | 120B | Differential Geometry II | 142 | Mathematical Modeling | 173B | Casualty Loss Models II |
| 61 | Discrete Structures | 121 | Toplogy | 151A | Numerical Analysis I | 174 | Mathematical Economics |
| 106 | History of Mathematics | 123 | Axiomatic Geometry | 151B | Numerical Analysis II | 180 | Graph Theory |
| 110A | Abstract Algebra I | 131A | Real Analysis I | 164 | Optimization | 184 | Combinatorics |
| 110B | Abstract Algebra II | 131B | Real Analysis II | 167 | Game Theory | 191 | Research Seminar |
| 110C | Abstract Algebra III | 131C | Real Analysis III | 170A | Probability I | 199 | Individual Research |

# APPENDIX
## A. COURSE NUMBERS

Table 5 shows the official course numbers and the associated course names used throughout the paper. We shortened the official course names wherever possible, to make the tables more visually appealing. For example, "Introduction to Fourier Analysis" is simply labelled "Fourier Analysis". Also, A, B, C are substituted with I, II, and III respectively. A short description of each course can be found in the UCLA general catalog [1]. A description of each major and its requirements can also be found at the same link. Lastly, in the general catalog, a list of official prerequisites can be found to validate that the sequences obtained by our rank aggregation methods are comparable with the official ones.